\documentclass[10pt,twocolumn,letterpaper]{article}

\usepackage[pagenumbers]{iccv}              %

\newcommand{\revision}[1]{\textcolor{black}{#1}}
\newcommand{\rev}[1]{\textcolor{black}{#1}}

\newcommand{\polys}{\mathbf{P}}

\newcommand{\loss}{\mathcal{L}}

\newcommand{\E}{\mathbb{E}}
\newcommand{\R}{\mathbb{R}}

\newcommand{\gf}{\psi}
\newcommand{\action}{\mathbf{A}}
\newcommand{\G}{\mathbb{G}}
\newcommand{\F}{{\mathcal{F}}}
\newcommand{\adj}{{\mathcal{A}}}
\newcommand{\sspace}{\mathcal{S}}

\newcommand{\funco}{O}
\newcommand{\grado}{\Phi}

\def\cf{\emph{c.f}.}

\def\etal{\emph{et al}. }

\def \gfpackplus {{GFPack++}}
\def \gfpack {{GFPack}}

\definecolor{iccvblue}{rgb}{0.21,0.49,0.74}
\usepackage[accsupp]{axessibility}  %
\usepackage[pagebackref,breaklinks,colorlinks,allcolors=iccvblue]{hyperref}

\hypersetup{
    pdftitle={GFPack++: Attention-Driven Gradient Fields for Optimizing 2D Irregular Packing},
    pdfauthor={Tianyang Xue, Lin Lu, Yang Liu, Mingdong Wu, Hao Dong, Yanbin Zhang, Renmin Han, Baoquan Chen}
}

\usepackage{booktabs} %
\usepackage{overpic}
\usepackage{graphicx}
\usepackage{amsmath}
\usepackage{array}
\usepackage{multirow}
\usepackage{textcomp}
\usepackage{diagbox}
\usepackage{algorithm}
\usepackage[noend]{algpseudocode}
\usepackage{amsmath}
\usepackage{enumitem}
\usepackage{wrapfig}
\usepackage{tablefootnote}
\usepackage{wrapfig}
\graphicspath{{paper/figures/}}

\def\confName{ICCV}
\def\confYear{2025}

\title{{\gfpackplus}: Attention-Driven Gradient Fields for Optimizing 2D Irregular Packing}

\author{
Tianyang Xue$^{1}$ \quad
Lin Lu$^{1\ast}$ \quad
Yang Liu$^{2}$ \quad
Mingdong Wu$^{3}$ \quad
Hao Dong$^{3}$ \\
Yanbin Zhang$^{1}$ \quad
Renmin Han$^{1}$ \quad
Baoquan Chen$^{3}$ \\
$^{1}$Shandong University \quad
$^{2}$Microsoft Research Asia \quad
$^{3}$Peking University \\
}

\begin{document}
\maketitle

\renewcommand{\thefootnote}{\fnsymbol{footnote}}
\footnotetext[1]{Corresponding author: llu@sdu.edu.cn}

\begin{abstract}

2D irregular packing is a classic combinatorial optimization problem with various applications, such as material utilization and texture atlas generation.
Due to its NP-hard nature, conventional numerical approaches typically encounter slow convergence and high computational costs. 
Previous research ({\gfpack}) introduced a generative method for gradient-based packing, providing early evidence of its feasibility but faced limitations such as insufficient rotation support, poor boundary adaptability, and high overlap ratios. 
In this paper, we propose {\gfpackplus}, a deeply investigated framework that adopts attention-based geometry and relation encoding, enabling more comprehensive modeling of complex packing relationships.
We further design a constrained gradient and a weighting function to enhance both the feasibility of the produced solutions and the learning effectiveness.
Experimental results on multiple datasets demonstrate that {\gfpackplus} achieves higher space utilization, supports continuous rotation, generalizes well to arbitrary boundaries, and infers orders of magnitude faster than previous approaches. 
Codes for this paper are at \href{https://github.com/TimHsue/GFPack-pp}{https://github.com/TimHsue/GFPack-pp}.

\end{abstract}

\section{Introduction}
\label{sec:introduction}

Irregular shape packing is a classic challenge in operation research, with significant implications for various applications such as texture atlas generation and material utilization in manufacturing~\cite{Wang2021}. Its NP-hard nature requires practical and effective packing algorithms to optimize space utilization.

Traditional numerical optimization methods are often slow and costly. Recently, learning-based methods have emerged as a promising alternative for irregular packing, mainly using reinforcement learning~\cite{Xu2023RLPack,Yang2023RLPack}. 
Another promising approach is the diffusion model, which learns geometric relationships from teacher packing examples and establishes a gradient field for all objects, guiding them to move simultaneously~\cite{Hosseini2022Puzzle, GFPack}. 
In particular, the work of \cite{GFPack}, namely {\gfpack}, achieves a notable trade-off between computational cost and space utilization, demonstrating scalability in terms of polygon count and generalizability to rectangular container boundaries.
However, it still faces unresolved issues, notably the lack of support for rotation, a relatively high occurrence of object collisions in results, and limited adaptability to arbitrary boundaries.

\begin{figure}[tb]
\includegraphics[width=\linewidth]{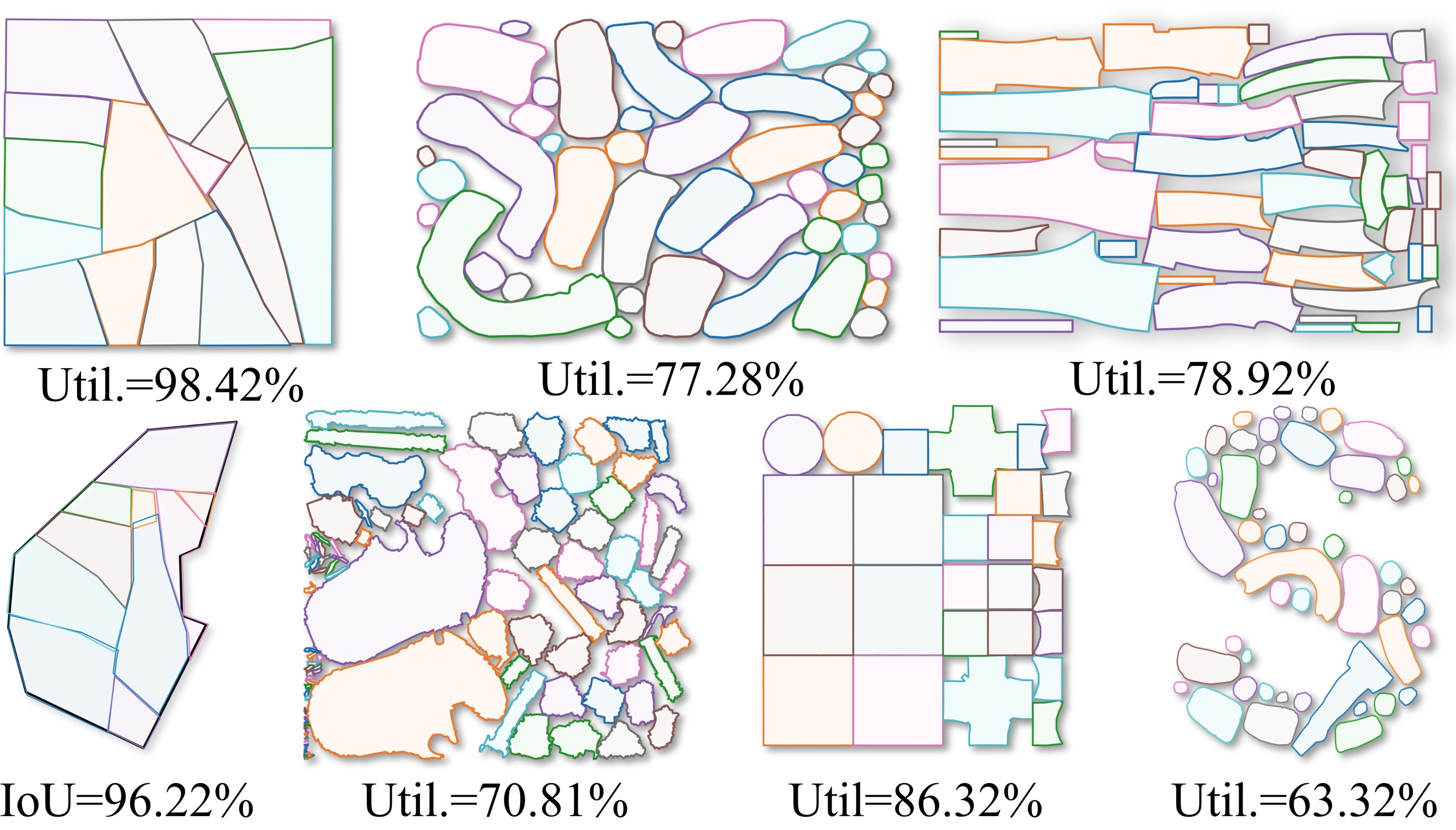}
\caption{The proposed attention-driven diffusion model for 2D irregular packing, \gfpackplus, supports
continuous rotation and accommodating arbitrary boundaries.}
\label{fig:teaser}    
\end{figure}

The main idea of the learning-based packing framework is to capture the geometric relationships between shapes, which are essential for the gradient computation. 
\revision{Such relationships in latent space encompass multiple facets: the \emph{spatial (position, orientation) relationship} of the shapes, the \emph{geometric features relationship} among them, and the \emph{container boundary conditions} that constrain them.
However, {\gfpack} falls short in fully representing these relationships. 
It solely focuses on the positional aspect, neglecting orientation, thus hindering rotational capabilities. Furthermore, it relies on PointNeXt~\cite{Qian2022PointNeXt} for geometric feature extraction, ignoring critical details like polygon edges and resulting in a loss of topological and local features. 
Thus, we argue that the full potential of gradient learning has not been thoroughly exploited in {\gfpack}.
}

In this study, we introduce two primary strategies to tackle the challenge of learning packing policies. 
Firstly, we use a graph to represent each object and encode it with an \emph{attention-based geometric encoder}, which can capture polygonal geometry information effectively.  
Secondly, we design an \emph{attention-based relation encoding} mechanism to encode intricate relationships both between polygons and among polygons and boundaries, \revision{eliminating the need to explicitly represent the relationships with a graph.}
This attention-based diffusion model enhances the understanding of how objects interact with each other and the boundaries, resulting in superior performance in both training and inference.
\rev{Moreover, ensuring fine-grained collision-free constraints is challenging for generative models.
We introduce a constraint gradient correction strategy driven by a collision potential field. 
The repulsion gradient between polygons is computed based on the Separating Axis Theorem and embedded into the reverse diffusion process.}
\revision{We also investigate the utilization distribution between the teacher and inference data and design a weighting function to prioritize tighter teacher data during training, confirmed to be effective.}

This paper makes the following contributions:
\begin{enumerate}[leftmargin=*]\setlength\itemsep{1mm}
    \item[-] We propose {\gfpackplus}, a novel packing framework that employs attention-based learning of gradient fields, supporting continuous rotation and enhancing generalization.

    \item[-]  \rev{Building on {\gfpack}, we use an attention-based encoder to capture spatial relationships (position, orientation), geometric features, and boundary conditions, facilitating effective gradient correction with a collision potential field.}  

    \item[-] \revision{{\gfpackplus} achieves superior irregular packing results, demonstrating higher space utilization over several widely used baselines, one order of magnitude faster than the previous diffusion-based method, coupled with promising generalization for arbitrary boundaries.}
\end{enumerate}

\section{Related Work}
\label{sec:related}

\paragraph{Traditional packing approaches}
Traditional methods usually involve placement optimization and sequence planning (\cf surveys~\cite{Bennell2009, Leao2020}). Most research focuses on placement optimization with heuristic rules (e.g., BL, BLF), NFP-based search~\cite{BennellNFP2008}, or mathematical models~\cite{Hopper2001}. Novel approaches include SDF-based placement~\cite{Pan2023SDFPack} and spectral domain correlation~\cite{CuiSpectralPack2023}, but they only support limited rotations. Sequence planning combines heuristics~\cite{svg_ref_1} with global search strategies like genetic algorithms\revision{~\cite{Junior2013}}, simulated annealing~\cite{Gomes2006SA}, beam search~\cite{Bennell2010beam}, \revision{or particle swarm optimization~\cite{Shalaby2013}}. However, these methods are computationally expensive and could improve space utilization by considering continuous rotation~\cite{Romanova2018}.

\paragraph{Reinforcement-learning-based packing}
Packing problems have attracted increasing interest in the field of machine learning, especially reinforcement learning (RL). RL methods can handle online 3D bin packing, where the object order is either fixed~\cite{Zhao2020RLPack, Zhao2022RLPack} or optimized~\cite{Zhang2021AttPack, Hu2020TAPNet, Xu2023RLPack}, also known as the transport-and-packing (TAP) problem. 
\revision{RL methods can also deal with irregular packing, where the objects have complex shapes and orientations~\cite{Huang2022RLPack, Zhao2022RLPack2}. 
A recent RL-based pipeline~\cite{Yang2023RLPack} integrates high-level object selection and low-level pose estimation and achieves a state-of-the-art packing ratio improvement of 5\% to 10\% over XAtlas~\cite{XAtlas} and NFP. 
However, RL-based packing methods share a commonality in considering object sequence and positioning separately, indicating room for improvement in achieving a better balance between space utilization and computational efficiency.
}

\begin{figure*}[tb]
\centering
\includegraphics[width=0.95\linewidth]{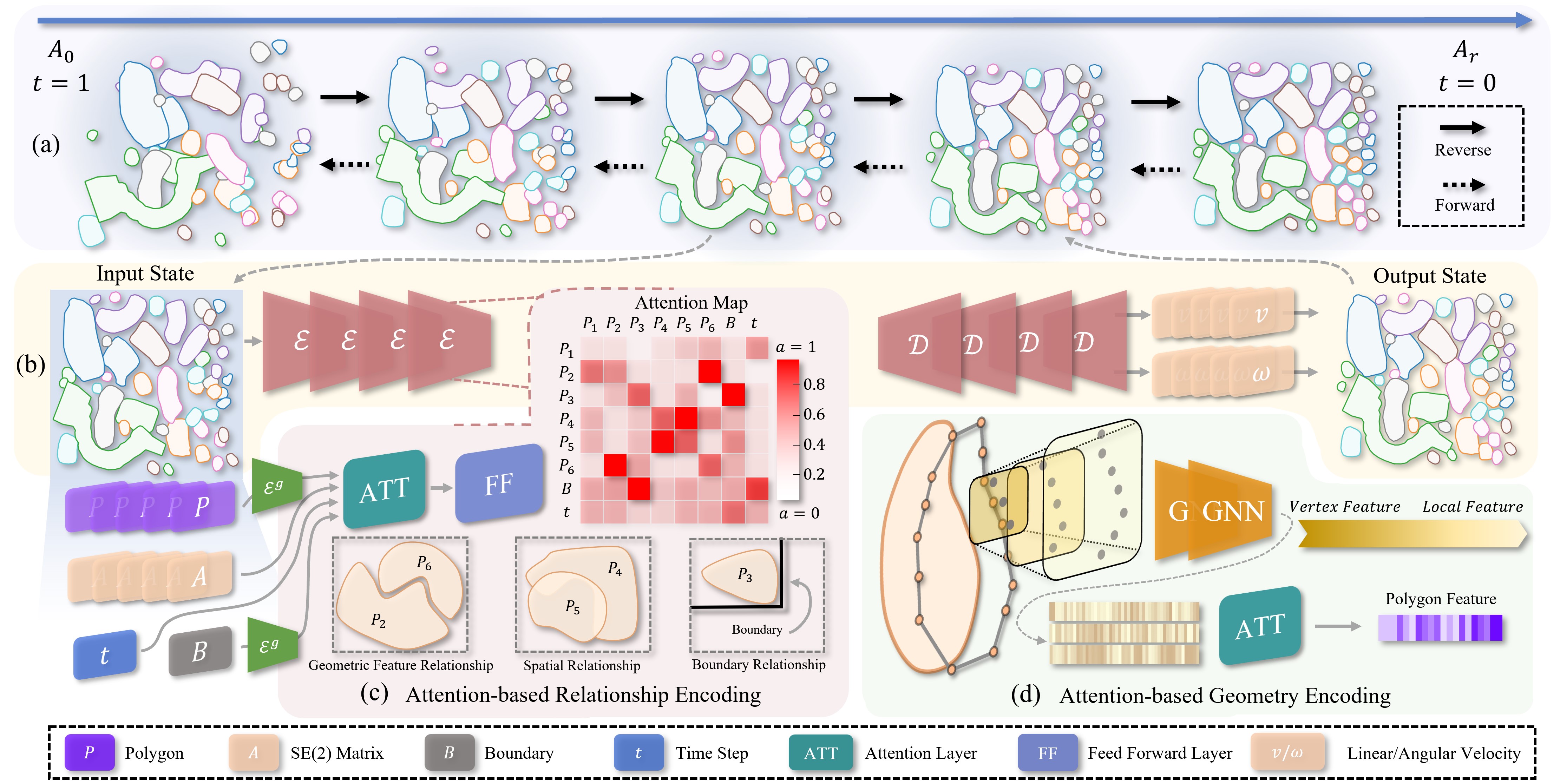}
\caption{
(a) Our approach is a diffusion-based generation method.
(b) We use a sequence-to-sequence model to encode and decode the state and velocities at each time step and generate the next state accordingly. 
Our model consists of two components: an attention-based relationship encoder (c) and an attention-based geometry encoder (d). The former encodes the geometric feature, spatial, and boundary relationships among the input variables using the attention mechanism. The latter computes the geometric features $\varepsilon^g$ using a multi-level GCN feature aggregation network and Attention Pooling.
}
\label{fig:overview}
\end{figure*}

\paragraph{Diffusion-based packing and arrangement}
\rev{
As a powerful generative model, diffusion models can effectively learn spatial relationships. Recent studies~\cite{Wu2022TarGF, Wei2023LEGO, Liu2022StructDif, FragmentDiff2024} show that object rearrangement tasks can implicitly learn object arrangements from training set, enabling object movement without explicit goals. }
However, object rearrangement does not require complex shape comparisons. Hosseini~\etal~\cite{Hosseini2022Puzzle} tackle this challenge by proposing a spatial puzzle solver using diffusion models. Nevertheless, puzzle solving focuses on pairwise object relationships as the ground truth solution exists, whereas packing problems lack such priors on the global optima, indicating a larger search space and necessitating more intricate consideration of geometric relationships.
{\gfpack}~\cite{GFPack} explores the use of diffusion models for 2D irregular packing problems. It learns geometric relationships from teacher examples and creates a gradient field for all objects, moving them simultaneously. 
While effective, {\gfpack} does not support rotation, exhibits frequent collisions, has limited adaptability to arbitrary boundaries, and is time-consuming. Our work addresses these issues by incorporating attention mechanisms into gradient field learning and designing an efficient learning framework.

\section{Overview}
\label{sec:overview}

\subsection{Problem Statement}

The 2D irregular packing problem aims to fit a set of 2D irregular polygons $\polys = \{P_i\}_{i=0}^n$ into a polygonal container $B$ with minimal wasted space. This problem can be formulated as an optimization problem, where the variables are the rigid transformations of each polygon, denoted as $\action = \{A_i\}_{i=0}^n$. 
\revision{The objective is to maximize the space utilization: $\mathit{u} = \bigcup_i \mathrm{Area}(A_i(P_i))/\mathrm{Area}(B)$, }subject to the constraints that each transformed polygon must be inside the container and not overlap with any other transformed polygon.
In this paper, we mainly set the container as an infinite strip that extends horizontally with a given height.
Our objective is to optimize the packing of $\polys$ to achieve the minimal length of the strip~\cite{Baker1980}, a practical consideration in most manufacturing scenarios.

\subsection{Learning Gradient Field with Attention}
Following \gfpack, we formulate the packing problem as a Markov Process. In this process, at each state, a velocity vector is generated for each polygon from the preceding state. This velocity includes both linear and angular components, as demonstrated in Fig. \ref{fig:overview} (a). Our goal is to learn a neural network that predicts the velocity at each timestep, using a teacher (training) set.

To generate the velocity vector, a state is encoded into a latent vector, which is then 'translated' by decoders into a velocity, as shown in Fig. \ref{fig:overview} (b). The accuracy of encoding implicit relationships within the state significantly influences efficiency. This is because, although the teacher data may not always be optimal, it often contains optimal local layouts. Therefore, an effective learning policy should be capable of discerning and learning such latent relationships in the teacher data. In {\gfpack}~\cite{GFPack}, a graph network determined by the distances between polygons is employed to facilitate state encoding. However, this approach overlooks the geometric relationships among the polygons. Moreover, the graph has to be rebuilt at each sampling step, leading to low efficiency.

To address these issues, we introduce two key designs. 
\revision{
We developed an attention-based relationship encoder that learns the encoding of multifaceted relationships between polygons and between polygons and boundaries, achieving high efficiency without explicitly updating the graph like {\gfpack}, as illustrated in Fig. \ref{fig:overview} (c).}
Additionally, we created a feature representation module to accurately capture the local and global geometric features of shapes, depicted in Fig. \ref{fig:overview} (d). This module ensures the comprehensive acquisition of geometric information, crucial for encoding the geometric relationships among shapes.
\revision{
Through the aforementioned designs, we have trained a score-based diffusion model to learn the gradient field for packing problems, accommodating continuous rotation and irregular boundaries. 
Accompanied by a weighting function that compels the model to focus on high-utilization data and utilizing an enhancement strategy for utilization, this results in the generation of high-utilization packing instances.
}

\section{Method}

We tackle the packing problem by encoding a state as polygon velocity, where the state $\sspace=\{\action, t, \polys , B\}$. 
Initially, polygons, composed of contour points present in both $\polys$ and $B$, are encoded as geometric features (Sec.~\ref{sec:gnn_feature}). 
Subsequently, these features, in conjunction with additional information $\action$ and $t$, are fed into a relation encoding module to generate velocity (Sec.~\ref{sec:att_relation}). 
\revision{
This entire process is trained through a score-based diffusion model (Sec.~\ref{sec:learninggf},~\ref{sec:weight}) and 
undergoes the utilization enhancement (Sec.~\ref{sec:post_opti}).
}

\subsection{Attention-based Geometry Encoding}
\label{sec:gnn_feature}

\begin{figure}[!tbp]
\centering
\includegraphics[width=0.8\linewidth]{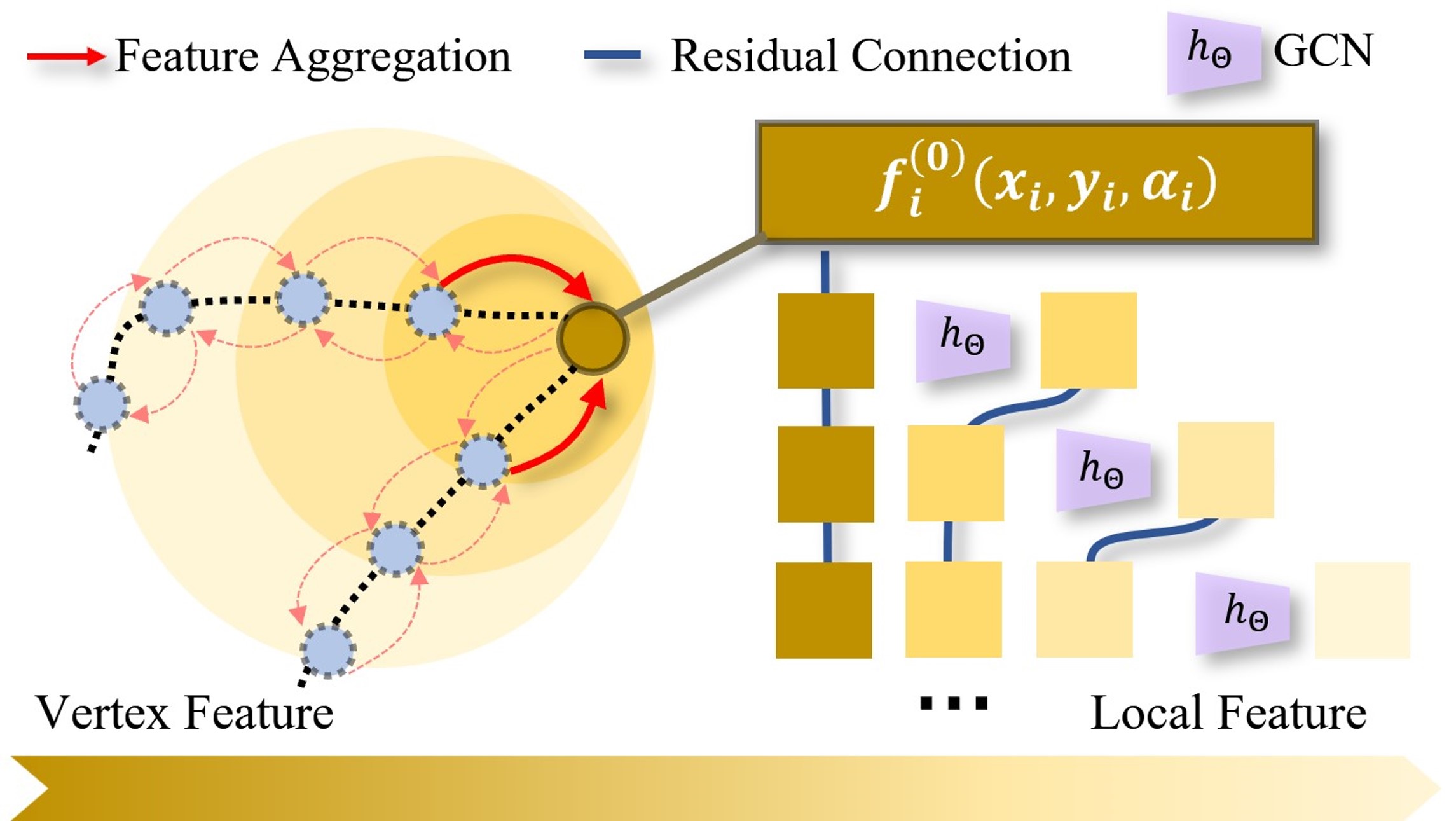}
\caption{Local Feature Representation. 
Our method leverages the information aggregation capability of GCNs combined with the multi-scale abilities of residual connections. 
It aggregates the geometric features of each point and its neighboring area with Eq.~\ref{eq:gnn}. }
\label{fig:gnn_feature}
\end{figure}

For a polygon contour point set $\mathbf{N} = \{N_i\}_{i=0}^{m}$, where $N_i \in \R^2$ and all vertices are given in order, we can readily construct a graph $\G=\{\mathbf{N}, \mathbf{E}=\{(i, i+1) | i \in [0,m]\}\}$.

Furthermore, for each vertex $N_i$, we compute its internal angle $\alpha$ and together with its coordinates, we form a set $f^{(0)}_i=(x_i, y_i, \alpha_i)$ as initial vertex feature.
The graph $\G$ and initial feature $f^{(0)}$ is then input into a GCN \cite{GCN2016}, and undergoes the following aggregation process:
\begin{equation}
\label{eq:gnn}
    h_\Theta(H^{(l)})=\sum_{(i,j)\in\mathbf{E}}\text{ReLU}(
    \hat{D}^{-\frac{1}{2}}
    \hat{\adj} \hat{D}^{-\frac{1}{2}}
    H^{(l)}W^{(l)}
    ),
\end{equation}
where $H^{(l)}$ is the features at layer $l$, $H^{(l)}=\{f^{(l)}\}$, $\hat{\adj}=\adj+I$ is the adjacency matrix with added self-connections, $\hat{D}$ is degree matrix of $\hat{\adj}$ and $W$ is a learnable weight matrix.
For each layer of GCN, we connect them as depicted in Fig. \ref{fig:gnn_feature}, employing a residual connection \cite{ResNet}:
\begin{equation}
\label{eq:residual}
    H^{(l+1)}=h_\phi( h_\theta(H^{(l)}), H^{(l)}).
\end{equation}
We achieved the local geometry encoding $H^{(L)}$ through the aggregation of $L$ layers.

\begin{figure}[!t]
\centering
\includegraphics[width=0.85\linewidth]{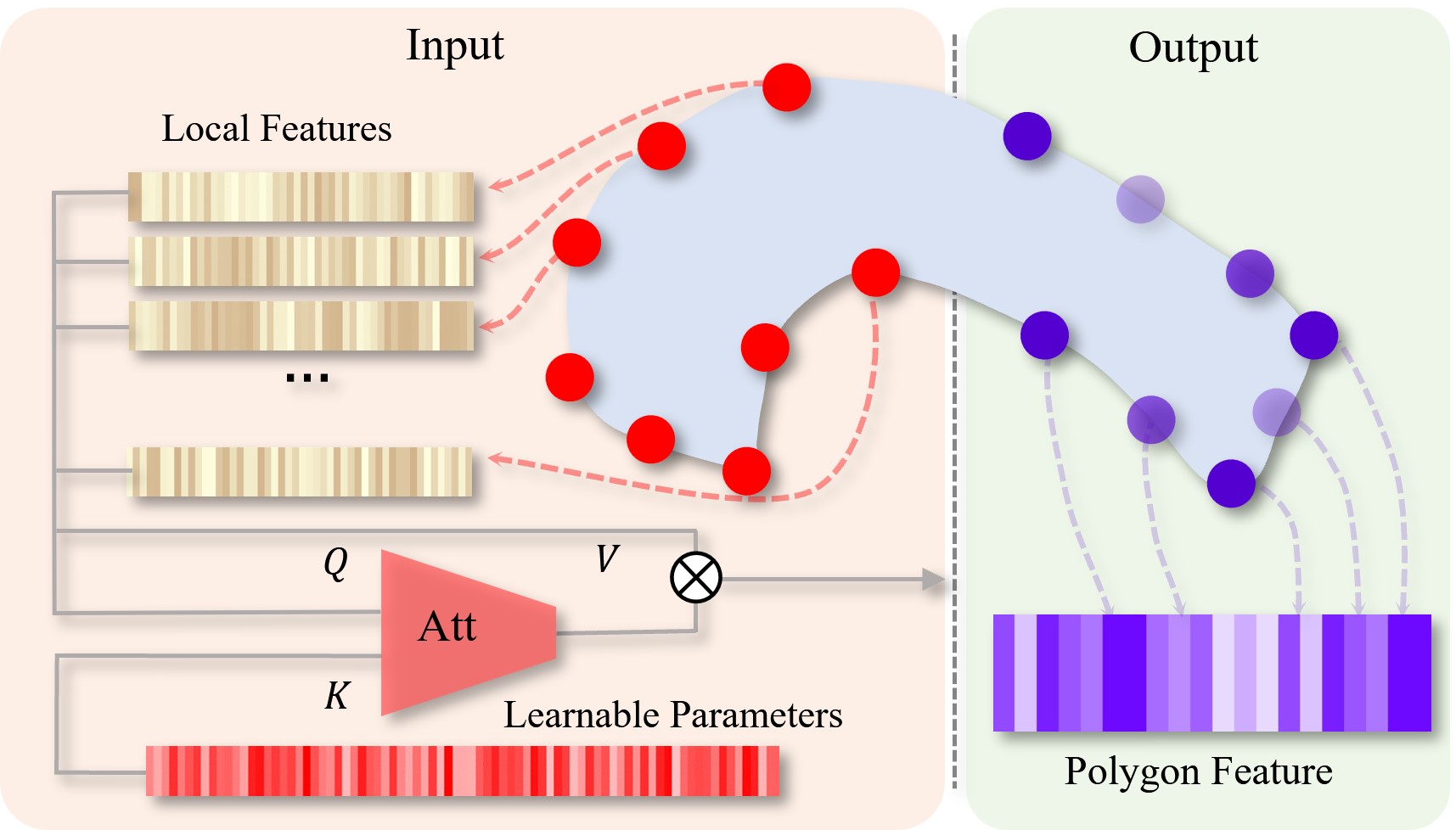}
\caption{Attention pooling for local features. 
For the local features of each point, we aggregate them into the shape features of the polygon using an attention mechanism.}
\label{fig:attpool}
\end{figure}

Inspired by \cite{attention2017, Velickovic2017GraphAN, Lee2019SelfAttentionGP}, we introduce an \textit{attention pooling} method to integrate local geometry features into a polygon geometry feature. 
The polygon geometry feature is computed as follows (Fig.~\ref{fig:attpool}):
\begin{equation}
\label{eq:att}
   \textrm{ATT}(Q, K, V)=\text{softmax}\Bigl(\frac{QK^T}{\sqrt{d_K}}\Bigr)V.
\end{equation}
Here, $Q$ (Query), $K$ (Key), and $V$ (Value) represent the key components of the attention mechanism, and $d_K$ is the dimension of $K$. In our design, $K=W_K$ is a learnable matrix, and $Q=V=H^{(L)}$.

Polygons $\polys$ and boundary $B$ are computed as geometric features $\F_{\polys}$ and $\F_{B}$ using the aforementioned method.

\subsection{Attention-based Relation Encoding}
\label{sec:att_relation}

Given the input state $\sspace_i=\{\action_i, t_i; \polys , B\}$ the output velocities $v$ is used to calculate the $\action_{i+1}$ in the next state with $\action_{i+1}=\action_{i} + v(\sspace_i)$.
We trained a Transformer \cite{attention2017} model to encode the state and generate the velocity, employing an attention mechanism to encode the geometric relationships mentioned above.

Initially, the state is encoded as a uniform-shaped latent feature vector with the following:
Polygons and boundary represented in contour are encoded through Sec.~\ref{sec:gnn_feature} as $\F_{\polys}\in \R^{|\polys|\times d_p}$ and $\F_{B}\in \R^{d_b}$.
SE(2) matrices are encoded through an MLP as $\F_{\action} \in R^{|\polys|\times d_a}$.

Time step $t$ is processed using Fourier Projection~\cite{Fourier2020}: $(t_x, t_y)=(\sin(2\pi w t), \cos(2\pi w t))$, where $w$ is initialized as $w \sim \mathcal{N}(0, I)$, $I\in \R^{d_t}$, $t_x$ and $t_y$ are feature vectors of shape $d_t$ calculated by $t$.
This projection transforms $t$ into a $2d_t$-dimensional feature vector.

We devise three attention-based relation encoding modules with Eq.~\ref{eq:att}.
The first is \textit{geometric feature relationships} module, derived from the geometric features among polygons, achieved through $Q=K=V=\F_{\polys}$. 
For encoding the \textit{spatial relationships}, we design $Q=V=(\F_{\polys}, \F_{\action}), K=\F_{\action}$.
Ultimately, we configure $Q=V=(\F_{\polys}, \F_{\action})$, $K=\F_B$ to process the \textit{boundary relationships}.
All these encoded features are fed into the next layer with a Feed Forward layer.

\subsection{Packing with Gradient Field}
\label{sec:learninggf}

We leverage the gradient field method from {\gfpack} to generate velocity with score-based diffusion. 
Specifically, we learn the log-density $\gf \approx \nabla_{\action}{\log{p_{opt}(\action)}}$ from sample distributions generated by a teacher packing algorithm.

We adopt gradient fields to global and local optimization phases through conditional time step $\{t_i \in [0, 1]\}_{i=0}^r$ as noise scale.
We incorporate Variance-Exploding Stochastic Differential Equation (VE-SDE) proposed by \cite{song2021scorebased} for noising control. 
\begin{equation}
\sigma(t) = \sigma_{\text{min}}\Bigl(\frac{\sigma_{\text{max}}}{\sigma_{\text{min}}}\Bigr)^t
\end{equation}
VE-SDE is capable of preserving the original data distribution, enhancing precision control for packing problems. 
This enables the calculation of the distribution at any time step $t$:
\begin{equation}
\label{eq:forward}
p(\action(t)|\mathcal{C}) = \int \mathcal{N}(\action(t)|\action(0), \sigma^2(t)\mathbf{I}) \cdot p(\action(0)|\mathcal{C}) \ d\action(0),
\end{equation}
where $\mathcal{C}=\{\polys, B\}$ and $p(\action(t)|\polys)$ represents the marginal distribution of $\action(t)$ conditional on $\mathcal{C}$.
The gradient field based on time step and boundary can be expressed as:
\begin{equation}
    \gf(\action, t;\mathcal{C}) \approx \nabla_{\action}{\log{p(\action(t)|\mathcal{C})}}.
\end{equation}
Conditioned on any $t \sim \mathcal{U}(0,1)$, we can train $\gf$ via Denoising Score Matching~(DSM)~\cite{denosingScoreMatching}:
\begin{equation}
   \loss(\gf) =  \E_{
   \action(t) \sim p(\action(t))
   }
   \left[\left\Vert\gf(\action, t; \mathcal{C})  - \frac{\action(0) - \action(t)}{\sigma(t)^2} \right\Vert_2^2 \right]. 
\label{eq:score_matching_loss}
\end{equation}

We acquire the final solution by solving the following reverse-time Random Stochastic Differential Equation (RSDE):
\begin{equation}
\label{eq:rsde}
    d\action = -g^2(t)\nabla_{\action}{\log{p(\action(t)|\mathcal{C})}}dt + g(t)\sqrt{dt}\epsilon,
\end{equation}
where $\sqrt{dt}\epsilon$ is the standard Wiener process with $\epsilon \sim \mathcal{N}(0, 1)$, $g(t)=\sigma(t) \sqrt{2\log{(\frac{\sigma_{\text{max}}}{\sigma_{\text{min}}})}}$ and $\nabla_{\action}{\log{p(\action(t)|\mathcal{C})}}$ is approximated by the learned gradient field $\gf(\action, t;\mathcal{C})$.
The solution path of the RSDE mentioned above can be exactly conceptualized as the velocities of the generation process $\{v(\sspace_i)\}_{i=1}^r$.

\subsection{Collision-free Constraint Gradient}
\label{sec:post_opti}

\rev{
Diffusion-based methods often struggle to enforce strict constraints, such as ensuring collision-free configurations.
To address this, we define $\funco(P_i, P_j)$ as the collision potential function between two polygons $P_i$ and $P_j$. 
The total collision potential for a single polygon $P_i$, denoted as $\grado_i$, is computed as the area-weighted average of its interactions with other polygons and the boundary:
}
\begin{equation}
\label{eq:overlap_grad}
\grado_i= \sum_{j>i}{\frac{{S_j \funco(P_i,P_j)}}{S_i + S_j}}
 - \sum_{j<i}{\frac{{S_j \funco(P_i,P_j)}}{S_i + S_j}}  + \funco(P_i,B),
\end{equation}
\rev{
where $S_i$ denotes the area of polygon $P_i$.  
}

\rev{
The positions of polygons are iteratively adjusted during the sampling process to minimize the collision potential.
This is achieved by incorporating the collision potential gradient into the gradient field, modifying Eq.~\ref{eq:rsde} as follows:
}
\begin{equation}
\label{eq:rsde_cor}
\begin{aligned}
    d\action = -g^2(t)(\nabla_{\action}{\log{p(\action(t)|\mathcal{C})}} - \nabla_{\action}\grado )dt + g(t)\sqrt{dt}\epsilon.
\end{aligned}
\end{equation}
\rev{
The gradient of collision potential function can be approximated using the Separating Axis Theorem~\cite{collisionDetection} as $\nabla_{\action}\funco(P_i, P_j) \approx \nu_i(i, j)$, where $\nu_i(i, j)$ represents the minimum displacement vector required to resolve the overlap between polygons $P_i$ and $P_j$.
}

\subsection{Training Optimization}
\label{sec:weight}
\revision{
We observed that the utilization rates of the training data influence the generation results, with higher utilization rates leading to better outcomes.
To prioritize high utilization data during learning, we apply a \textit{sigmoid} weighting to the loss function in Eq.~\ref{eq:score_matching_loss}.}
\revision{$\lambda(\action, \mathcal{C})$ directs the network's focus towards optimal layouts:}
\begin{equation}
\label{eq:weight}
    \lambda(\action, \mathcal{C}) = \text{sigmoid}({
        \frac{
            u(\action, \mathcal{C}) - \mathit{U}_{avg}
        } {
            \mathit{U}_{max} - \mathit{U}_{min}
        } \times 10
    }),
\end{equation}
\revision{where $\mathit{U}_{min|avg|max}$ represents the Min$|$Avg$|$Max utilization ratio across the entire teacher set, respectively.}

\revision{
This weighting function effectively balances the quality and diversity of the training data. 
We conducted ablation studies on this function, with details provided in the supplementary materials.
}

\section{Implementation}

\begin{table*}[!ht]
\centering
\setlength{\abovecaptionskip}{5pt}
\caption{
Statistics of packing ratios (Min{\textbar}Avg{\textbar}Max) and time consumption for the average packing results. All of these results were optimized with enhancement algorithm. 
}

  \resizebox{\textwidth}{!}{
    \begin{tabular}{cccccc}
      \toprule
      \textbf{Dataset} & \textbf{XAtlas}                & \textbf{NFP}                   & \textbf{SVGnest}               & \textbf{{\gfpackplus} ($b=128$)} & \textbf{{\gfpackplus} ($b=512$)}        \\ \midrule
      Garment          & 62.26\,{\textbar}\,69.53\,{\textbar}\,74.70\,{\textbar}\,1.01 s  & 61.49\,{\textbar}\,65.49\,{\textbar}\,67.94\,{\textbar}\,6.13 s  & 69.38\,{\textbar}\,72.75\,{\textbar}\,75.13\,{\textbar}\,80.21 s & 69.37\,{\textbar}\,74.22\,{\textbar}\,77.83\,{\textbar}\,3.25 s    & \textbf{74.15\,{\textbar}\,77.04\,{\textbar}\,80.62\,}{\textbar}\,10.5 s  \\
      Dental           & 63.18\,{\textbar}\,70.86\,{\textbar}\,73.59\,{\textbar}\,1.25 s  & 60.42\,{\textbar}\,66.30\,{\textbar}\,68.13\,{\textbar}\,8.22 s  & 67.22\,{\textbar}\,73.64\,{\textbar}\,76.49\,{\textbar}\,96.73 s & 70.57\,{\textbar}\,75.21\,{\textbar}\,78.67\,{\textbar}\,4.28 s    & \textbf{74.06\,{\textbar}\,77.53\,{\textbar}\,79.79}\,{\textbar}\,14.5 s  \\
      Puzzle (square)  & 56.16\,{\textbar}\,66.48\,{\textbar}\,76.15\,{\textbar}\,1.21 s & 56.90\,{\textbar}\,62.53\,{\textbar}\,68.13\,{\textbar}\,4.52 s  & 61.00\,{\textbar}\,67.22\,{\textbar}\,73.21\,{\textbar}\,42.55 s & 82.22\,{\textbar}\,93.99\,{\textbar}\,98.64\,{\textbar}\,6.23 s   & \textbf{84.80\,{\textbar}\,95.12\,{\textbar}\,98.74}\,{\textbar}\,32.1 s  \\
      Atlas (building) & 50.56\,{\textbar}\,71.21\,{\textbar}\,87.62\,{\textbar}\,0.96 s & 41.97\,{\textbar}\,67.32\,{\textbar}\,83.31\,{\textbar}\,2.76 s  & 44.97\,{\textbar}\,74.51\,{\textbar}\,90.54\,{\textbar}\,52.43 s & 58.96\,{\textbar}\,78.78\,{\textbar}\,98.47\,{\textbar}\,8.12 s    & \textbf{66.03\,{\textbar}\,80.16\,{\textbar}\,98.88}\,{\textbar}\,25.3 s  \\
      Atlas (object)   & 38.58\,{\textbar}\,60.98\,{\textbar}\,76.09\,{\textbar}\,1.43 s & 32.18\,{\textbar}\,57.75\,{\textbar}\,83.12\,{\textbar}\,45.3 s & 41.53\,{\textbar}\,63.87\,{\textbar}\,83.12\,{\textbar}\,402.2 s & 31.76\,{\textbar}\,65.11\,{\textbar}\,86.08\,{\textbar}\,15.5 s    & \textbf{41.95\,{\textbar}\,67.47\,{\textbar}\,86.81}\,{\textbar}\,65.4 s \\ \bottomrule
    \end{tabular}
  }
\label{tab:efficiency_result}
\end{table*}

\subsection{Datasets and Teacher Algorithm}
\label{sec:data_teacher}

\begin{figure}[h]
    \centering
    \includegraphics[width=1.0\linewidth]{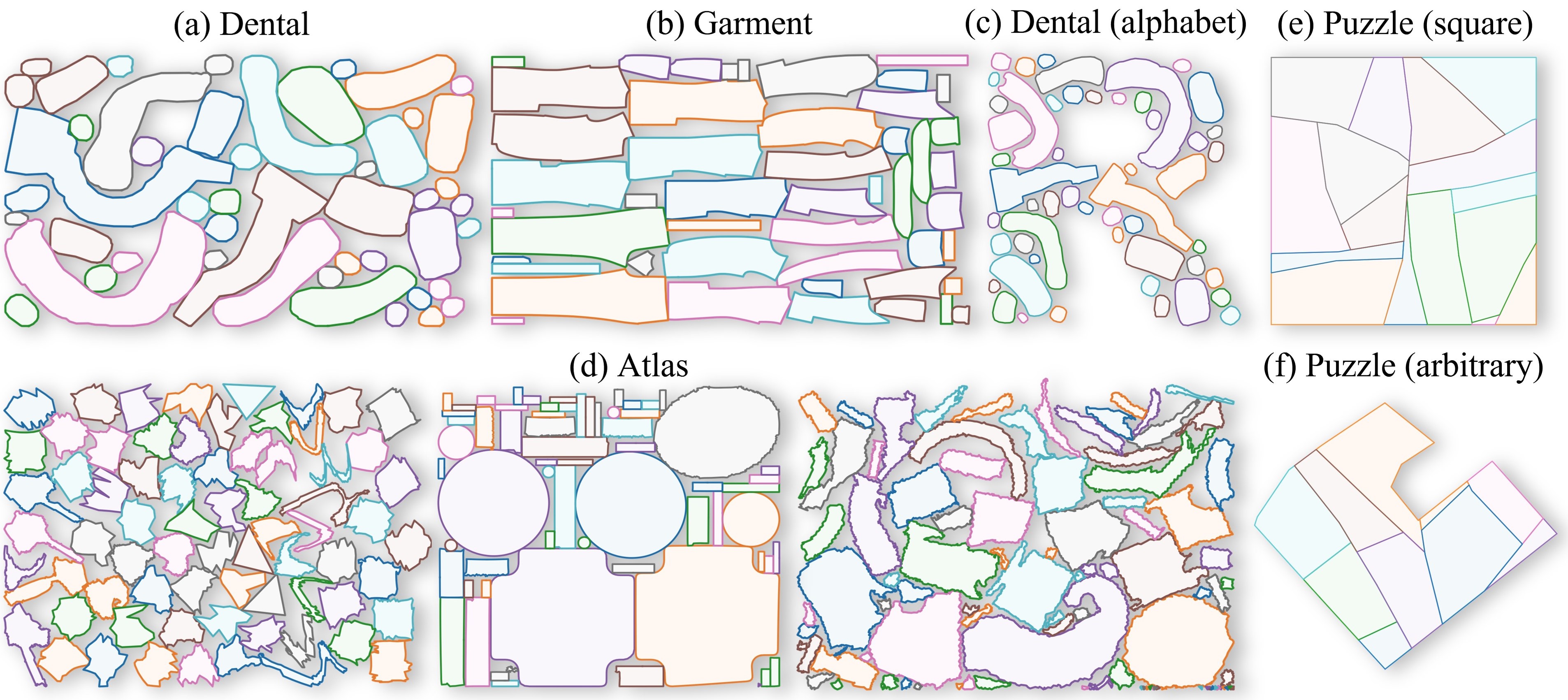}
    \caption{Samples from the teacher datasets.}
    \label{fig:dental_garment_dataset}
\end{figure}

We validated our algorithm using \revision{six} datasets (Fig.~\ref{fig:dental_garment_dataset}).
The Garment and Dental datasets stem from real manufacturing scenarios \cite{GFPack}. 
We introduced a dataset, named Dental (alphabet), generated by packing dental data according to alphabetical boundaries.
\revision{We also employed Atlas datasets from \cite{Yang2023RLPack}, which contain the UV patches of 3D models.}

\revision{For these four datasets, we utilized SVGnest~\cite{SVGnest} as the teacher algorithm to generate exemplar packing results (training data distributions are provided in the supplementary materials). SVGnest employs a packing position selection based on NFP and a genetic algorithm for sequence and discrete angle selection \cite{svg_ref_1, svg_ref_2}. In the rotational learning study, we set the range of discrete angle selection to 32 and applied random rotations to the polygons before inputting them into the teacher algorithm. This strategy helps the model learn the distribution of arbitrary rotation angles. The SVGNest's population size and number of iterations were both set to 8. We ran the algorithm 10 times on the same set of polygon combinations and selected the best result.
}

\revision{
We introduced two new puzzle datasets, Puzzle (square) and Puzzle (arbitrary), to assess the method's generalizability across various polygon shapes and boundaries.
Both datasets feature randomly generated polygon shapes and irregular boundaries, creating "ground-truth" packing solutions with 100\% utilization. 
Puzzle (square) comprises 16 polygon fragments per item, while Puzzle (arbitrary) consists of 10 fragments. 
Additionally, Puzzle (arbitrary) includes 8192 boundaries randomly selected from pieces generated in Puzzle (square).
The generation algorithm for each data item involves randomly dividing a polygon boundary into multiple parts, ensuring no intersection.}

The training data for each dataset comprises approximately 100,000 teacher items of random, non-colliding polygon combinations. 
\rev{Detailed polygon information and utilization ratio distributions for each dataset are provided in the Supplementary Sec.~\ref{supp:sec:dataset}.}

\subsection{Experimental Setup}
\label{sec:exp_set}

In {\gfpackplus}, we employed a four-layer ($L=4$) Graph Convolutional Network (GCN) in Sec.~\ref{sec:gnn_feature} for local feature extraction, with attention pooling, ultimately generating feature vectors $F_{\polys}$ and $F_B$ with $d_p=64$, $d_b=128$.
The SE(2) matrices $\action$ and time step $t$ are encoded to $F_{\action}$ with $d_a=64$, $d_t=64$.

The feature vectors $\F_{\polys}$ and $\F_{\action}$ are concatenated together to form a 128-dimensional feature vector.
Along with $\F_{\action}$ and $\F_t$, they are fed into the transformer structure with 8-layer encoders and 8-layer decoders (Sec.~\ref{sec:att_relation}). 
Through the attention mechanism, relationship information is encoded and the velocity is outputted.

At each training step, a random $t$ is selected from the range $[0.01, 1]$. The corresponding perturbed $A'$ with a noise level at $t$ is then input into the network, calculated based on the formulas in Sec.~\ref{sec:learninggf} with $\sigma_{min}=0.1$ and $\sigma_{max}=1000$, aligned with packing results. Additionally, for computational convenience, the polygon's translations and rotations are represented as $(x, y, 1000\times\cos(\theta), 1000\times\sin(\theta))$ before being input into the network. This ensures uniform data scales in both rotation and translation spaces.
Finally, we compute the \textit{Mean Squared Error} (MSE) between the network output and the perturbation, as described in Eq.~\ref{eq:score_matching_loss}, optimizing using the AdamW optimizer~\cite{loshchilov2018decoupled} at a learning rate of $2\times10^{-4}$.

During generation, for a given polygon set, we initialize $A(t_0) \sim \mathcal{N}(0, \sigma(t_0))$ with $t_0=1$. 
We then sample $r=128$ steps using the Euler-Maruyama method~\cite{kloeden2011numerical} to generate $A_r$ through RSDE in Eq.~\ref{eq:rsde}.
\rev{
In the later stages of generation ($r > 96$), the constraint gradient described in Sec.~\ref{sec:post_opti} is applied to resolve collisions.
The constraint gradient is computed in C++ with parallelization for efficiency.
After generation, this algorithm can run additional iterations until reaching a step limit (100 steps) or convergence separately.
To further improve utilization, we implement a rapid gap elimination method inspired by \cite{CuiSpectralPack2023}. Binary search is used during the above iterations to calculate the minimum distance polygons can move in both $-x$ and $-y$ directions, which is then applied to reduce gaps between polygons. 
Together, this method and the constraint gradient are referred to as utilization enhancement.
The impact of this enhancement could be found in Supplementary Sec.~\ref{supp:sec:impact}.
}
For each packing task, {\gfpackplus} generates multiple solutions in parallel (batch size $b$) and selects the collision-free solution with the highest utilization as the final result.

\section{Results}
\label{sec:results}

\subsection{Experimental Analysis} 
\label{sec:util}

\paragraph{Baselines.}
We applied strip packing to the Garment, Dental, and Puzzle datasets. The strip heights were randomly set for Garment, 1200 for Dental, and 2000 for Puzzle. 
{\gfpackplus} was trained and tested on datasets containing 48 polygons for Dental and Garment, and 16 polygons for Puzzle (square). 
The Atlas datasets were tested without fixed boundaries, with the number of polygons varying from 5 to 200.
Following model training, we randomly generated new test sets, each comprising 128 instances, to evaluate the algorithms.
The statistics are summarized in Table~\ref{tab:efficiency_result}. 
Among these, XAtlas~\cite{XAtlas} and NFP are non-sequential optimization algorithms known for their efficiency in single-step position selection, making them relatively fast. 
\rev{
XAtlas was configured in brute-force mode and adapted to support strip packing, with details in the Supplementary Sec.~\ref{supp:sec:baseline}. 
}
SVGnest~\cite{SVGnest}, which also serves as our teacher algorithm, combines sequential optimization and achieves superior space utilization at the expense of increased time consumption. 
{\gfpackplus} demonstrates superior space utilization by 4\%-32\% compared to the three baseline algorithms, attributed in part to our learning of optimal layouts and the ability to explore a continuous optimal distribution of rotation. 
\rev{These results are further visualized in Supplementary Secs.~\ref{supp:sec:visualization} and~\ref{supp:sec:comparison}.}

\begin{figure}[ht]
\centering
\includegraphics[width=.9\linewidth]{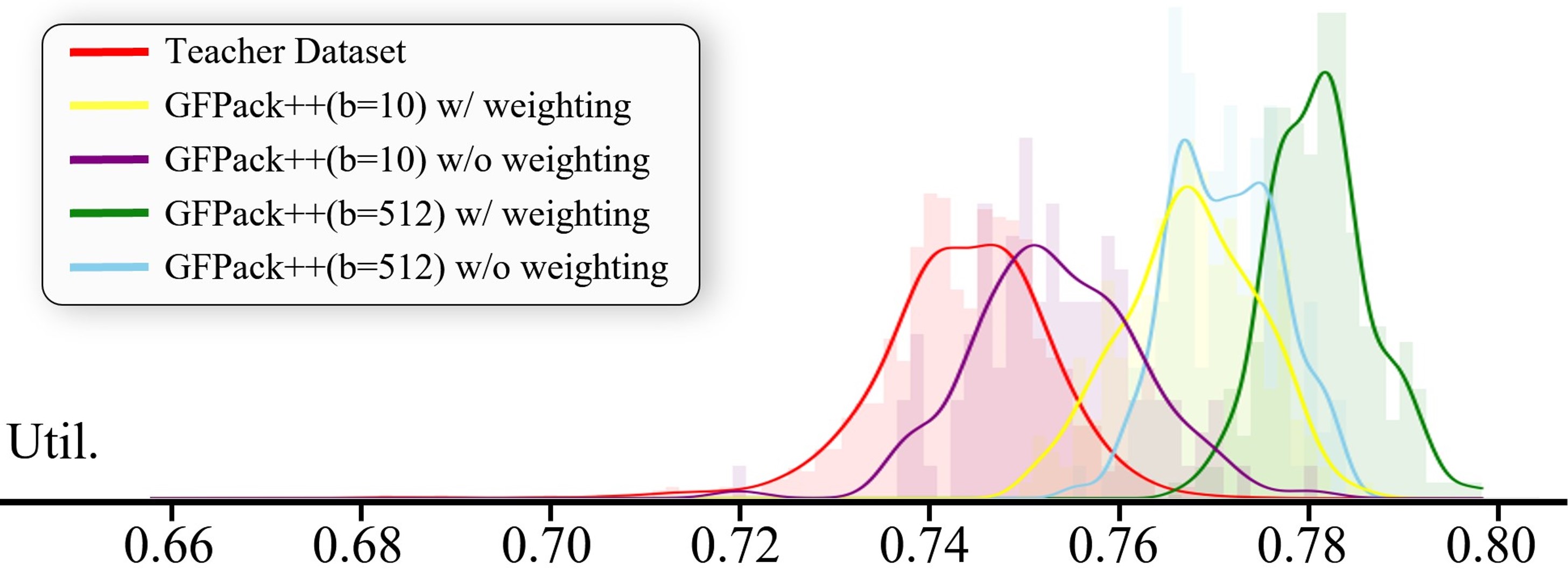}
\caption{\revision{Utilization distribution shift for the dental data. The figure plots the histogram of utilization ratios of 1000 random data.}}
\label{fig:shift}
\end{figure}

\vspace{-12pt}
\paragraph{Utilization distribution shift.}
\revision{
We observed that the utilization ratio of teacher data is enhanced by {\gfpackplus}, and is further improved by our weighting function introduced in Sec.~\ref{sec:weight}. The experimental setup was as follows: we selected $1000$ data randomly from the teacher dataset and conducted inference by {\gfpackplus} trained with and without the weighting function. The utilization distribution of the dental data, depicted in Fig.~\ref{fig:shift}, shows a shift towards greater utilization when the weighting function is applied, indicating that training data with higher utilization are more effective to the training. Moreover, employing strategies that allow for the simultaneous generation of multiple solutions on the batch size \( b \) enhances utilization. 
We chose two $b$ values, as the teacher data is chosen from 10 samples (Sec.~\ref{sec:data_teacher}) and {\gfpackplus} inference is performed with $b=512$.
}

\begin{figure}[ht]
\centering
\includegraphics[width=1.0\linewidth]{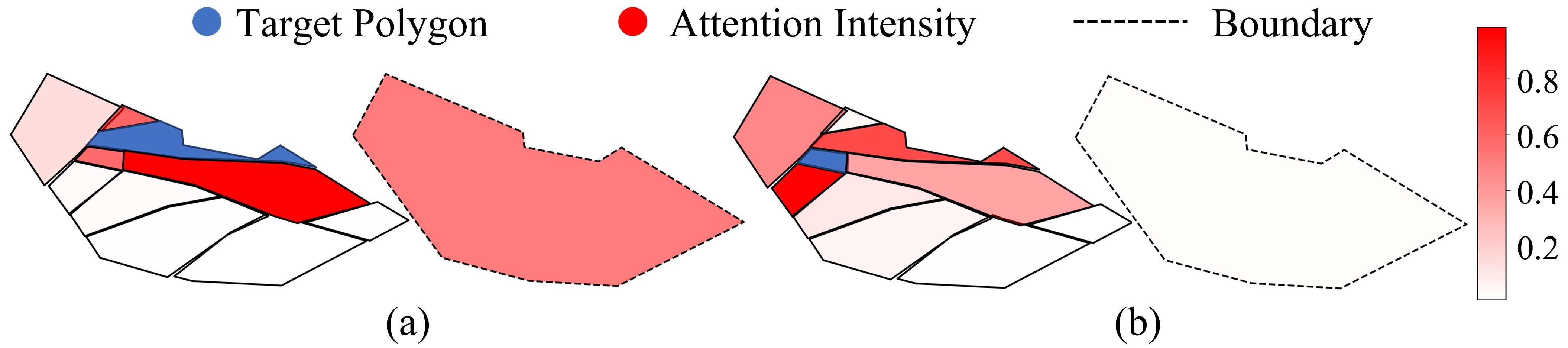}
\caption{Visualization of the attention intensity of the 7th-layer decoder at $t=0.2$ using data from the Puzzle (arbitrary) dataset. Attention intensity of each polygon is normalized by $\frac{a_i-a_{min}}{a_{max} - a_{min}}$. }
\label{fig:att_vis}
\end{figure}

\vspace{-12pt}
\paragraph{Attention-based relation encoding visualization.} To better understand the relations represented by the attention mechanism, we visualize an attention weight from our model.
Fig.~\ref{fig:att_vis} illustrates the attention intensity at $t=0.2$, closing to the convergence.
Polygons close to the boundary exhibit stronger attention towards the boundary, whereas polygons fully enveloped by surrounding polygons demonstrate lower boundary attention intensity. 
Additionally, polygons display near-zero attention intensity towards those polygons with which they are not adjacent.

\subsection{{\gfpackplus} vs. {\gfpack}}

\revision{
In Table~\ref{tab:gfp_vs_gfpp}, {\gfpackplus} is compared with {\gfpack}, showcasing notable enhancements in utilization, overlap reduction, and computational efficiency. {\gfpackplus} exhibits a 5\% to 7\% increase in utilization when both methods employ utilization enhancement techniques, and it operates an order of magnitude faster than {\gfpack}. 
}
\begin{table}[ht]
\centering
\setlength{\abovecaptionskip}{5pt}
\caption{Comparison of {\gfpack} and {\gfpackplus} on Garment and Dental. 'E' indicates enhancement of model outputs, 'R' denotes training with arbitrary rotations. 'Util.' refers to the average spatial utilization, 'Over.' is the average overlap of generated outcomes as a percentage of total polygon area, and 'Time' represents the average generation time.} 
\scalebox{0.8}{
\begin{tabular}{clrrr}
\toprule
\textbf{Dataset}         & \textbf{Algo.}     & \textbf{Util.(\%)} & \textbf{Over.(\%)} & \textbf{Time(s)} \\ \midrule
\multirow{4}{*}{Garment} & {\gfpack}          & 69.82     & 0.85      & 81.2    \\
                         & {\gfpack} (E)       & 72.17     & 0.23      & 124.0     \\
                         & {\gfpackplus}      & 74.25     & 0.08      & 7.9    \\
                         & {\gfpackplus} (E)   & \textbf{77.04}     & \textbf{0.00}      & 10.5    \\ \midrule
\multirow{6}{*}{Dental}  & {\gfpack}          & 66.59     & 1.43      & 80.4    \\
                         & {\gfpackplus}      & 69.31     & 0.06      & 8.0    \\
                         & {\gfpack} (E)       & 70.11     & 0.78      & 132.1     \\
                         & {\gfpackplus} (E)   & 73.93     & \textbf{0.00}      & 12.1    \\
                         & {\gfpackplus} (R)   & 75.11     & 0.09      & 8.4    \\
                         & {\gfpackplus} (R+E) & \textbf{77.53}     & \textbf{0.00}      & 14.5    \\ \bottomrule
\end{tabular}
}
\label{tab:gfp_vs_gfpp}
\end{table}

\revision{
Regarding the overlapping area ratio, {\gfpackplus} consistently reduces it to under 0.1\% of the total area, effectively eliminated by the enhancement algorithm. In contrast, {\gfpack} not only generates larger overlaps but also requires more time for their elimination. The feasibility of solutions, measured as the average number of feasible solutions per batch—was also evaluated. Despite the use of enhancement algorithms, {\gfpack} yielded only 64.2\% feasible solutions in the garment dataset and 34.2\% in the dental dataset. Meanwhile, {\gfpackplus} achieved 100\%, 98.4\%, and 96.3\% average rates of feasible solutions across the garment, dental, and rotation-inclusive dental datasets, respectively. Significantly, {\gfpack} was unable to produce feasible solutions for the rotation-inclusive dental dataset, with the learning policy struggling to determine reasonable velocities amidst the greatly increased complexity of shape relationships.
}

\revision{
To further clarify the disparities between {\gfpack} and {\gfpackplus}, we conducted ablation experiments on the proposed encoders. 
Table~\ref{tab:ablation} shows that an accurate representation of polygonal geometric features helps to reduce overlap areas.
}
\rev{
Detailed training information for {\gfpack} and {\gfpackplus} is provided in Supplementary Sec.~\ref{supp:sec:training}.
}
\begin{table}[ht]
\setlength{\abovecaptionskip}{5pt}
\centering
\caption{Performance comparison of {\gfpackplus} and {\gfpack} using different geometric encoders (first column) and relational encoders (second column) on the dental dataset. The results do not include enhancement.}
\scalebox{0.9}{
\begin{tabular}{llcc}
\toprule
\textbf{Geometric Enc.}  & \textbf{Relation Enc.}                   & \textbf{Over.}  & \textbf{Util.}   \\ \midrule
{\gfpack}               & {\gfpack}         & 13.2\%  & -\\
{\gfpackplus}           & {\gfpack}         & 8.38\%  & -\\
{\gfpack}               & {\gfpackplus}     & 1.89\%  & 72.84\%\\
{\gfpackplus} (AvgPool) & {\gfpackplus}     & 0.85\%  & 73.11\%\\
{\gfpackplus}           & {\gfpackplus}     & \textbf{0.08\%}  & \textbf{74.25\%}\\ \bottomrule
\end{tabular}
}
\label{tab:ablation}
\end{table}

\subsection{Generalization and Scalability} 
\label{sec:generalization}
\begin{figure}[ht]
\centering
\includegraphics[width=0.85\linewidth, trim=0 2mm 0 16mm, clip]{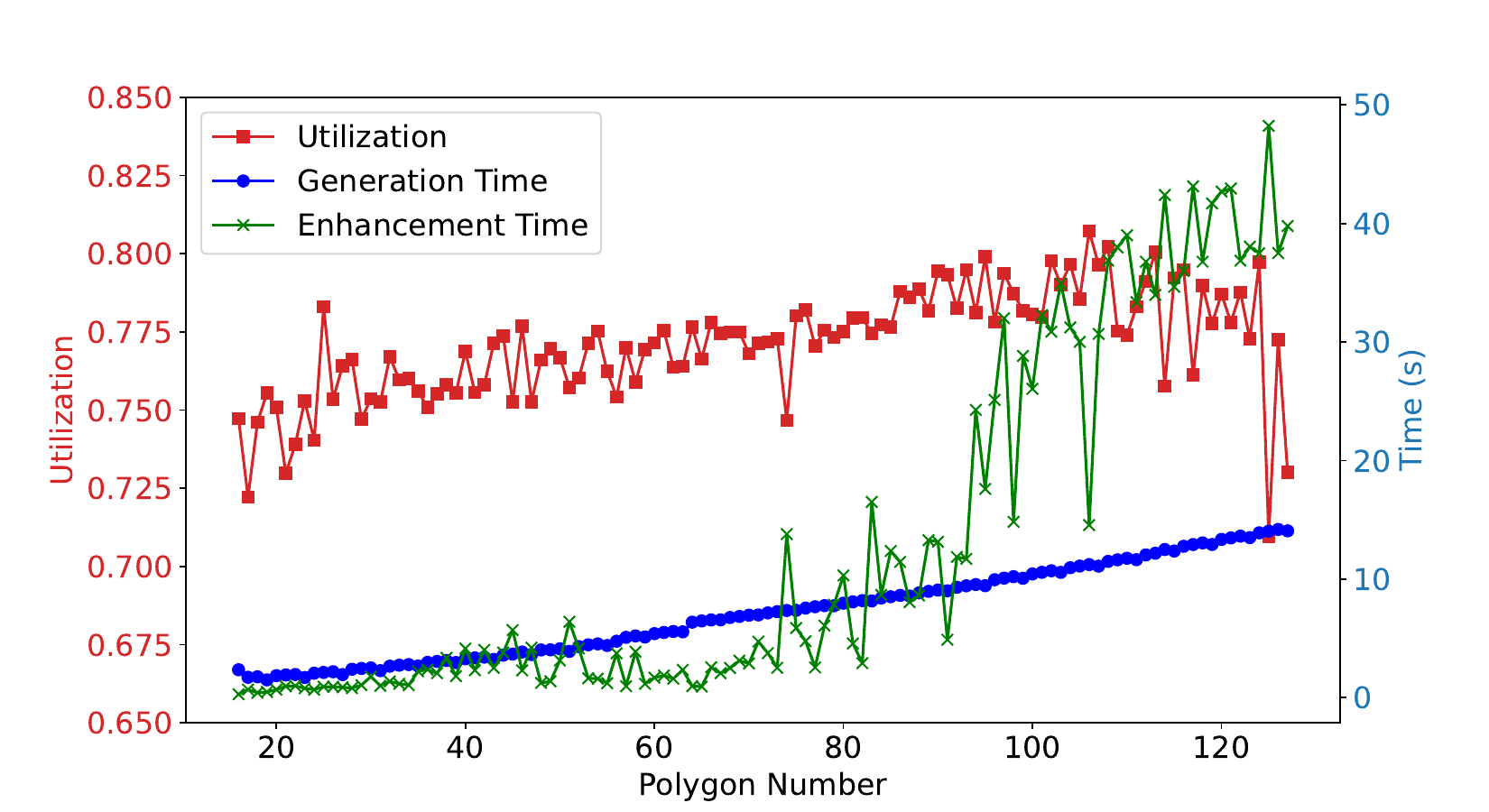}

\caption{Polygon number scalability of {\gfpackplus} (16 to 128).}
\label{fig:number}
\end{figure}

\paragraph{Number of polygons}
\revision{We tested the scalability of {\gfpackplus} to the number of polygons in the Dental dataset, by
training the model with a training set 24 to 64 polygons and validating it on a test set with 16 to 128 polygons. As shown in Fig.~\ref{fig:number}, %
our method maintains a low level of time consumption while avoiding significant decreases in space utilization when the number of polygons increases.}

\vspace{-10pt}
\begin{table}[ht]
\centering
\caption{Generalization test on Dental (alphabet).}
\scalebox{0.9}{
\begin{tabular}{lcc}
\toprule
\textbf{Batch Size}        & \textbf{Valid Solutions}  & \textbf{Time} \\ \midrule
{\gfpackplus} ($b=128$)    & 38.53\%             & 4.21 s    \\ 
{\gfpackplus} ($b=512$)    & 67.54\%             & 7.95 s    \\ \bottomrule
\end{tabular}
}
\label{tab:dental_alpha}
\end{table}

\vspace{-15pt}

\paragraph{Boundary and polygon shape variation}

\revision{
The generalization capabilities of {\gfpackplus} were assessed for polygon shapes and arbitrary boundaries using the Dental and Puzzle datasets. For the Dental dataset, 128 groups of valid data were generated, and the model attained a packing success ratio of 67.54\% with a batch size of 512, as detailed in Table~\ref{tab:dental_alpha}. In the case of the Puzzle dataset, 128 new boundaries with corresponding test data were created, approaching packing as a puzzle-solving task without enhancements. Performance was gauged using the Intersection over Union (IoU), defined as $IoU = \frac{B \cap (\cup \mathbf{P})}{B \cup (\cup \mathbf{P})}$, which quantifies the proportion of non-overlapping area within the boundary. This metric, akin to the overlap score in \cite{Hosseini2022Puzzle}, underscores the model's proficiency. Table~\ref{tab:puzzle} demonstrates {\gfpackplus}'s high IoU, underscoring its applicability in puzzle-solving contexts, while supplementary materials provide some visual results.
}

\begin{table}[t]
\centering
\setlength{\abovecaptionskip}{5pt} 
\caption{Generalization test on Puzzle (arbitrary).}
\scalebox{0.85}{
\begin{tabular}{lcc}
\toprule
\textbf{Batch Size}        & \textbf{IoU (Min$|$Avg$|$Max)}                & \textbf{Time} \\ \midrule
{\gfpackplus} ($b=128$)    & 80.39 \% $|$ 93.75 \% $|$ 96.80 \%   & 3.27 s    \\ 
{\gfpackplus} ($b=512$)    & 82.43 \% $|$ 94.91 \% $|$ 98.86 \%   & 4.23 s    \\ \bottomrule
\end{tabular}
}
\label{tab:puzzle}

\end{table}

\vspace{-5pt}
\paragraph{Comparison with \cite{Yang2023RLPack}}
\revision{
Table~\ref{tab:learn2pack_bui} shows that our method achieves results comparable to those in \cite{Yang2023RLPack}, which is designed for Atlas data. 
Our training data are generated by SVGnest, and we believe using the data from \cite{Yang2023RLPack} (unavailable for these experiments) could further improve our model. Training exclusively on the general dataset resulted in only a slight performance decrease. Refer to supplementary materials for visualizations.
}

\begin{table}[ht]
\centering
\setlength{\abovecaptionskip}{5pt}
\caption{Comparison of packing ratios (Min|Avg|Max) between {\gfpackplus} and \cite{Yang2023RLPack}. {\gfpackplus}* means our method trained on the general dataset.}
\scalebox{0.9}{
\begin{tabular}{lcc}
\toprule
\textbf{Algo.}                                          & \textbf{Building}                    & \textbf{Object}   \\ \midrule
\cite{Yang2023RLPack}                                   & \textbf{68.3} $|$ \textbf{82.7} $|$ 98.0   & 37.7 $|$ \textbf{68.7} $|$ 86.2       \\
{\gfpackplus}                                           & 66.0 $|$ 80.2 $|$ \textbf{98.9}   & \textbf{42.0} $|$ 67.5 $|$ \textbf{86.8}            \\
{\gfpackplus}*                                 & 61.5 $|$ 78.4 $|$ 97.4   & 36.2 $|$ 66.2 $|$ 83.9                 \\ \bottomrule
\end{tabular}
}
\label{tab:learn2pack_bui}

\end{table}

\section{Conclusions and Perspectives}

We present {\gfpackplus}, a novel packing framework utilizing attention-based learning of gradient fields.
 {\gfpackplus} supports continuous rotation, substantially improving generalization capabilities. It outperforms baseline algorithms in terms of space utilization and delivers efficiency that is an order of magnitude faster than prior diffusion-based methods. 
The source code and data are available through our \href{https://github.com/TimHsue/GFPack-pp}{project repository}.

\vspace{-5pt}
\paragraph{Future work}

\revision{
Inspired by the observed utilization distribution shift,
we speculate that the incorporation of {\gfpackplus}'s generation results into the teacher data could further enhance the learning.
We are also interested in developing a unified model for general 2D irregular packing, 
with challenges potentially arising in generating sufficiently diverse geometric features concisely and handling domain discrepancies.}
Furthermore, {\gfpackplus} demonstrates potential in puzzle-solving, prompting an exploration of its adaptation to fragment reassembling problems~\cite{Huang2006,sellan2022breaking}. %
Lastly, we plan to extend the proposed diffusion model to address 3D irregular packing problems.

\section{Acknowledgments}
We thank all reviewers for their valuable comments and constructive suggestions. 
This work is supported in part by grants from National Key Research and Development Program of China (Grant No. 2024YFB3309500) and National Natural Science Foundation of China (Grant Nos. 62472258, U23A20312).

{
    \small
    \bibliographystyle{ieeenat_fullname}
    \bibliography{paper/main}
}

\clearpage
\section*{Supplementary Material}
\setcounter{section}{0}
\setcounter{figure}{0}
\setcounter{table}{0}
\renewcommand{\thesection}{S\arabic{section}}
\renewcommand{\thesubsection}{\thesection.\arabic{subsection}}
\renewcommand{\thefigure}{S\arabic{figure}}
\renewcommand{\thetable}{S\arabic{table}}
\renewcommand{\theHsection}{supp.\arabic{section}}
\renewcommand{\theHfigure}{supp.\arabic{figure}}
\renewcommand{\theHtable}{supp.\arabic{table}}

The supplementary materials include visualizations and detailed information, such as the complexity of polygons in each dataset and statistics from both training and testing.

\section{Dataset Complexity}
\label{supp:sec:dataset}
\label{supp:sec:polygons}

\begin{figure*}[!t]
\centering
\includegraphics[width=1.0\linewidth]{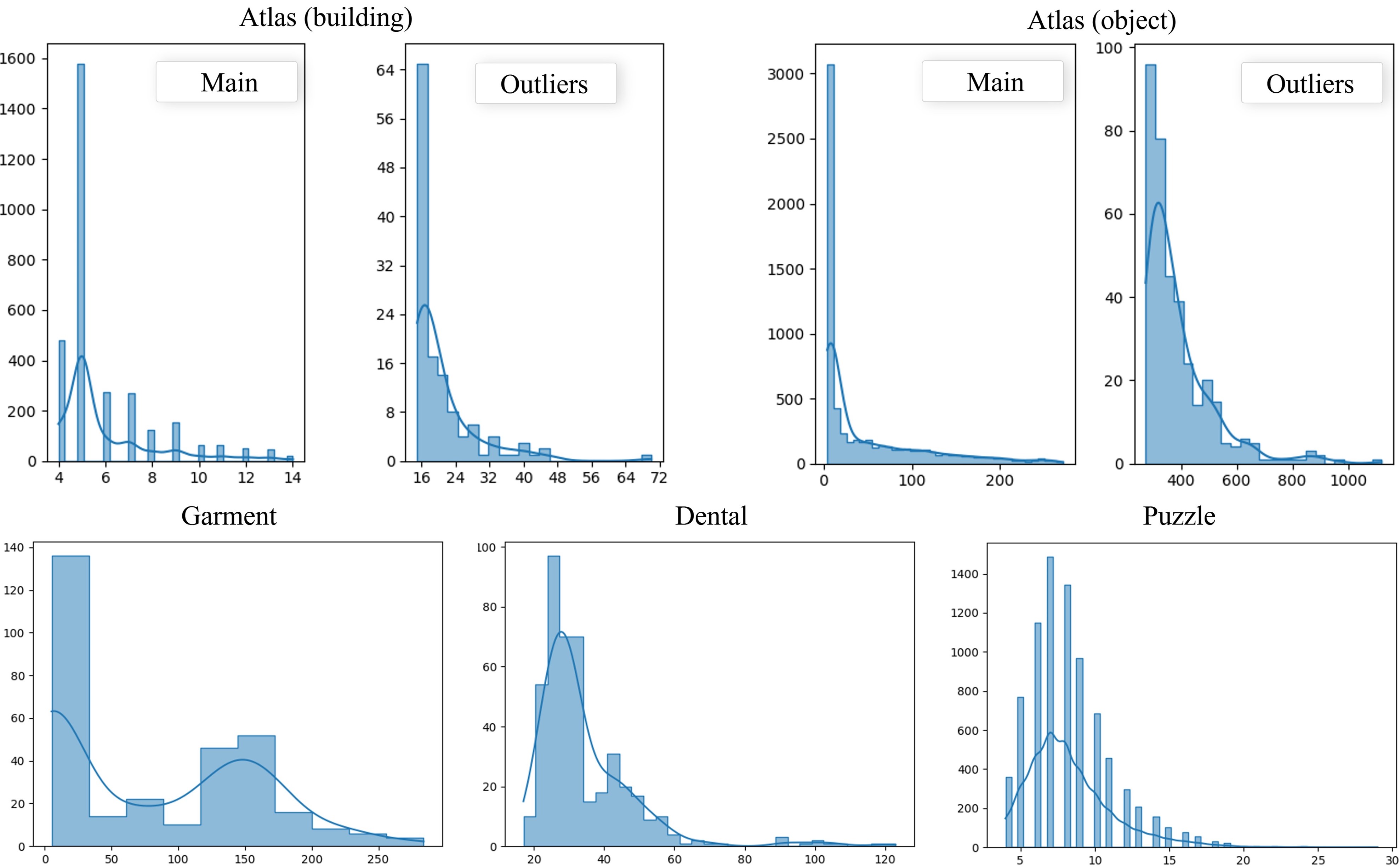}
\caption{The statistics of polygon vertex counts. 
The horizontal axis represents the number of vertices in a polygon, while the vertical axis indicates the frequency of these vertex counts. 
Due to the wide distribution of vertex counts in the Atlas dataset, we have divided it into two parts: the main distribution and the outliers.
}
\label{supp:fig:vertex_counts}
\end{figure*}

The distribution of our training data is described in Table~\ref{supp:tab:traning_dist}. 
The distribution of vertex numbers in the polygons of each dataset is presented in Table~\ref{supp:tab:vertex_counts} and Fig.~\ref{supp:fig:vertex_counts}. 
Every polygon in the Puzzle dataset is unique.
\begin{table}[ht]
\centering
\caption{Teacher dataset utilization distributions.}
\begin{tabular}{cccc}
\toprule
\textbf{Teacher dataset}        & \textbf{Min}        & \textbf{Avg}         & \textbf{Max}   \\ \midrule
Garment               & 63.02\%    & 72.39\%     & 78.90\%    \\ 
Dental                & 67.33\%    & 74.16\%     & 77.36\%    \\ 
Atlas (building)      & 39.92\%    & 74.55\%     & 95.74\%    \\ 
Atlas (object)        & 24.69\%    & 63.25\%     & 85.18\%    \\ 
Atlas (general)       & 40.76\%    & 67.54\%     & 92.41\%     \\ \bottomrule
\end{tabular}
\label{supp:tab:traning_dist}
\end{table}

Therefore, we randomly sampled 1000 polygons from the Puzzle dataset to calculate the vertex counts.
The number of polygons in each dataset is displayed in Table~\ref{supp:tab:poly_counts}. 
Samples of the polygons are shown in Fig.~\ref{fig:dental_garment_dataset}.

\begin{table}[ht]
\centering
\caption{Polygon vertex counts.}
\begin{tabular}{crrrr}
\toprule
\textbf{Dataset}               & \textbf{Min}      & \textbf{Avg}       & \textbf{Max}     & \textbf{Std} \\ \midrule
Garment               & 5        & 81.57     & 284     & 76.83 \\ 
Dental                & 17       & 34.47     & 123     & 14.50     \\ 
Puzzle                & 4        & 8.25      & 29      & 2.86   \\
Atlas (building)      & 4        & 6.52      & 70      & 3.87    \\ 
Atlas (object)        & 4        & 67.41     & 1120    & 102.38  \\  \bottomrule
\end{tabular}
\label{supp:tab:vertex_counts}
\end{table}

\begin{table}[ht]
\centering
\caption{Polygon counts in each dataset.}
\begin{tabular}{lr}
\toprule
\textbf{Dataset}               & \textbf{Polygon Count}  \\ \midrule
Garment               & 313            \\ 
Dental                & 440            \\ 
Atlas (building)      & 3262           \\ 
Atlas (object)        & 6764           \\ \bottomrule
\end{tabular}
\label{supp:tab:poly_counts}
\end{table}

\section{Impact of Enhancement}
\label{supp:sec:impact}

The enhancement process, which integrates the constrained gradient optimization (as described in Sec.~\ref{sec:post_opti} of the main text) and gap elimination (detailed in Sec.~\ref{sec:exp_set} of the main text), is employed to resolve minor overlaps and eliminate gaps.
The impact of enhancement on other algorithms is not significant (about 1\%, see Table~\ref{supp:tab:post_opt}), as they do not contain overlaps, have relatively small gaps, and offer little room for local optimization.
\begin{table}[!ht]
\centering
\caption{Average improvements achieved through enhancement.}
\scalebox{0.75}{
\begin{tabular}{lcccccc}
\toprule
\multirow{2}{*}{\textbf{Dataset}} 
                         & \multicolumn{2}{c}{\textbf{\gfpackplus}}   & \multicolumn{2}{c}{\textbf{XAtlas}}                        & \multicolumn{2}{c}{\textbf{SVGnest}}                     \\
                         & \multicolumn{1}{c}{\textbf{Bef.}} & \multicolumn{1}{c}{\textbf{Aft.}} & \multicolumn{1}{c}{\textbf{Bef.}} & \multicolumn{1}{c}{\textbf{Aft.}} & \multicolumn{1}{c}{\textbf{Bef.}} & \multicolumn{1}{c}{\textbf{Aft.}}\\ \midrule
Garment                  & 74.25\%                & 77.04\%     & 68.90\%                & 69.53\%                & 70.98\%                & 72.75\%     \\
Dental                   & 75.11\%                & 77.53\%     & 69.88\%                & 70.86\%                & 72.55\%                & 73.64\%     \\ \bottomrule
\end{tabular}
}
\label{supp:tab:post_opt}
\end{table}

\section{Training Settings}
\label{supp:sec:training}
\label{supp:sec:setting}

We conducted our training and testing on a Linux server equipped with 4 NVIDIA RTX 4090 GPUs and an Intel(R) Xeon(R) Platinum 8163 CPU.
{\gfpackplus} converged in 50 hours. {\gfpack} converged after 168 hours of training. 
We continued training for an additional 48 hours after convergence for both methods. 

The factor 10 in Eq.~\ref{eq:weight} of the main text amplifies the sigmoid input, effectively increasing the weight contrast between high- and low-utilization samples. This helps the network better focus on high-quality data. Empirically, we found that larger factors tend to suppress low-quality samples too aggressively, while smaller ones fail to sufficiently distinguish useful data. A value of 10 achieves a good balance and yields the best overall performance.

We quadrupled the number of parameters in {\gfpack} from 10 M to match {\gfpackplus} at 40 M.
This increase resulted in excessive VRAM usage and extended convergence times, with the model not converging within 240 hours.
Despite this, {\gfpack} still failed to produce collision-free results on datasets with rotations.
We did not further increase the number of training parameters since the training time was already too long.

\section{Baseline Algorithms Configuration}
\label{supp:sec:baseline}
We modified the code of XAtlas to enable its application to non-atlas packing problems. 
Key modifications include: 
(a) the introduction of a packing height constraint to facilitate strip packing operations; 
(b) adjustments to padding, alignment, and scaling methods to prevent distortion. 
Detailed XAtlas settings are provided here:
\begin{verbatim}
options.pack.height = height;
options.pack.rotation = true; 
options.pack.rotateChartsToAxis = align;
options.pack.bruteForce = true;
options.pack.padding = 0.5f;
options.pack.bilinear = true;
options.pack.blockAlign = false;
options.scale = 1.0f;
\end{verbatim}

In this context, \texttt{height} specifies the strip height, with a setting of -1 indicating unfixed boundaries.
XAtlas supports rotations of 0 and 90 degrees when both brute force and rotation are enabled. 
The align parameter determines whether polygons are aligned to their bounding boxes. 
We have disabled this setting for the Dental dataset, while it is enabled for other datasets.

Additionally, for SVGNest, we have enhanced its parallel efficiency to improve its performance within the same time.

\section{Validation Visualizations}
\label{supp:sec:visualization}
\begin{figure*}
\centering
\includegraphics[width=1.0\linewidth]{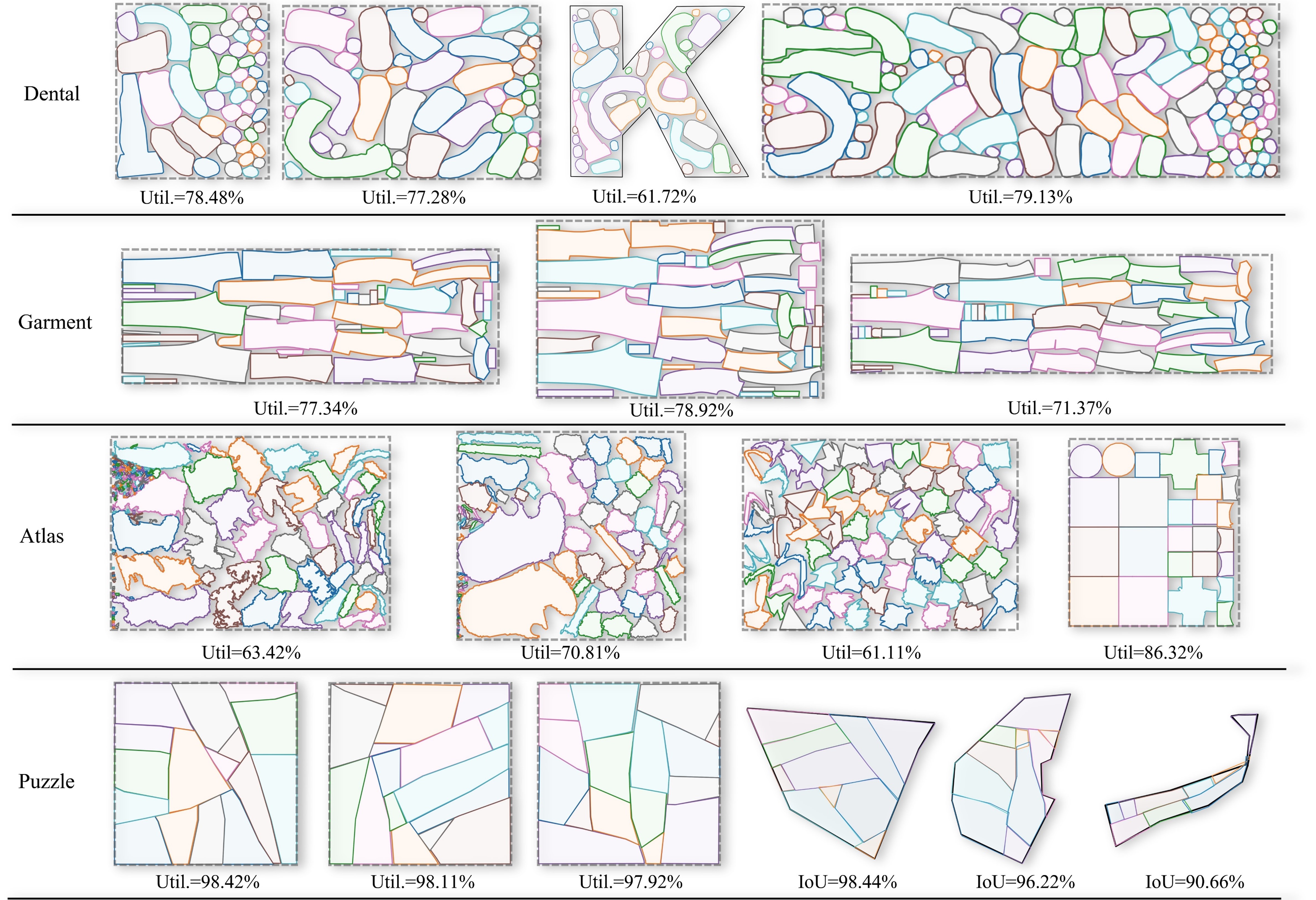}
\caption{Some results from {\gfpackplus} across different datasets. Note that the utilization enhancement only applies to regular boundaries, so small overlaps may be observed in the puzzle (arbitrary) results. 
}
\label{supp:fig:results}
\end{figure*}

We present visualizations of validation results (Sec.~\ref{sec:util} of the main text) generated by {\gfpackplus} across various datasets (Fig.~\ref{supp:fig:results}).

\noindent
\begin{minipage}[t]{0.72\linewidth}
\vspace{0pt}
We also observed failure cases in two typical scenarios: 
(1) \textbf{highly complex boundaries or polygon shapes}, which may lead to conflicting gradient fields and hinder convergence; 
and (2) \textbf{extreme overcrowding}, where the number of polygons far exceeds the layout capacity, leading to inevitable overlaps. 
Two such examples are shown in the right inset.
\end{minipage}\hfill
\begin{minipage}[t]{0.25\linewidth}
\vspace{0pt}
\centering
\includegraphics[width=\linewidth]{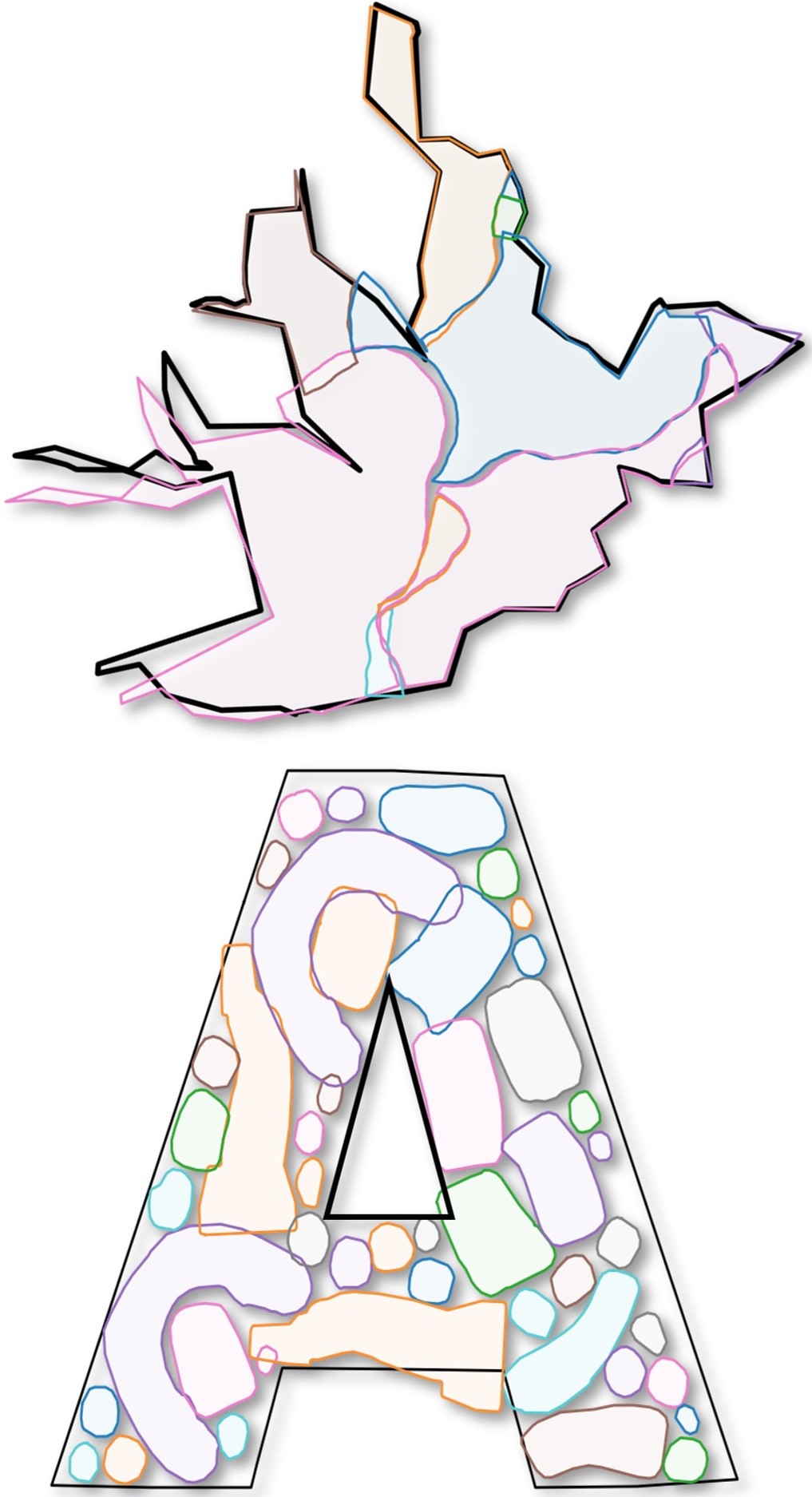}
\end{minipage}
\par

\section{Comparative Visualizations}
\label{supp:sec:comparison}
\begin{figure*}
\centering
\includegraphics[width=1.0\linewidth]{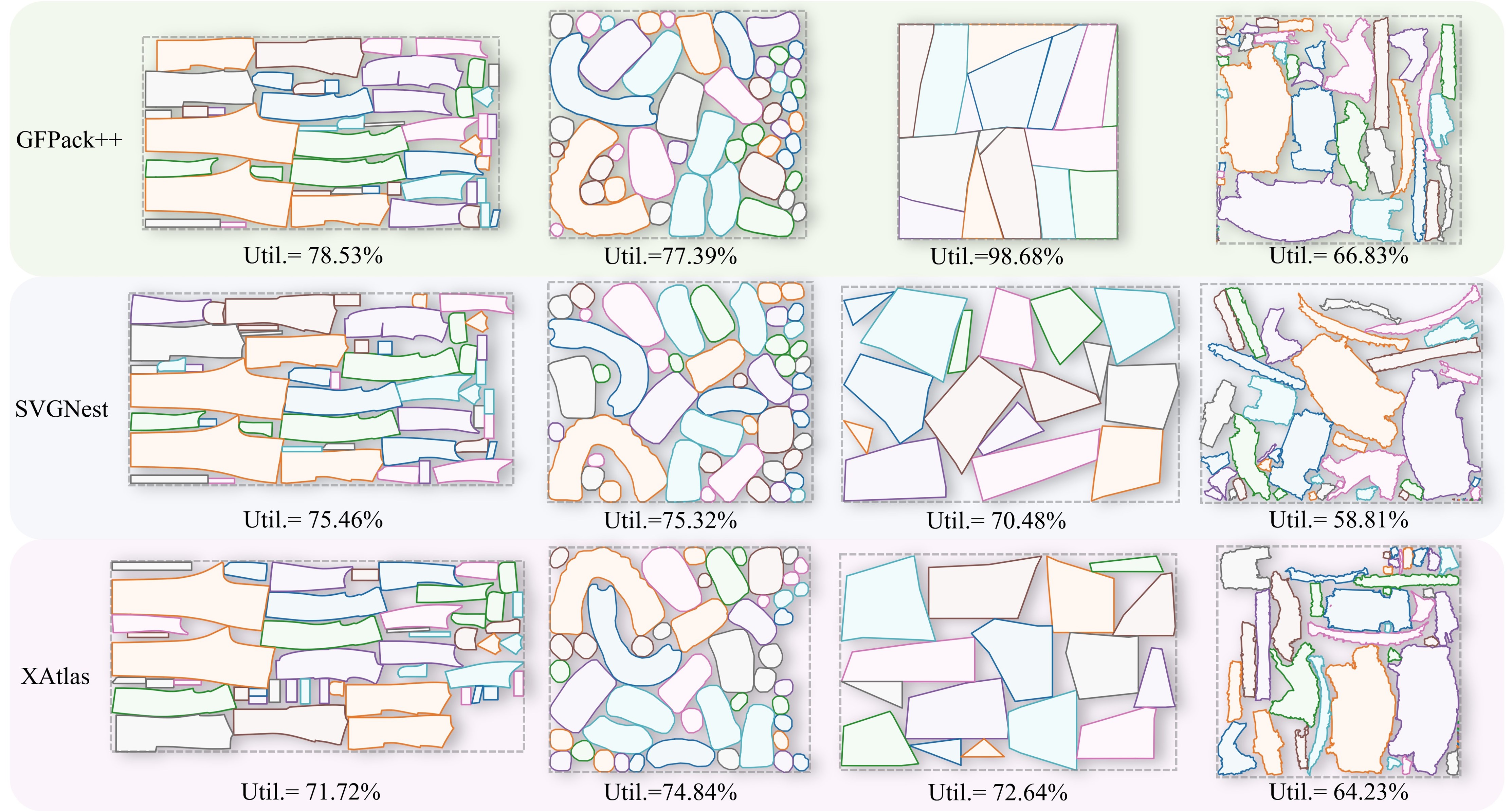}
\caption{Comparative results from {\gfpackplus}, SVGnest, and XAtlas applied to Garment, Dental, Puzzle, and Atlas datasets. 
In the lower left corner of the Garment dataset, a shared similar local layout pattern between {\gfpackplus} and SVGnest can be observed.
For the Puzzle dataset, {\gfpackplus} outperforms the others, nearly achieving a global optimum due to its continuous rotation solving.
}
\label{supp:fig:compare2XAtlas}
\end{figure*}

\begin{figure*}
\centering
\includegraphics[width=1.0\linewidth]{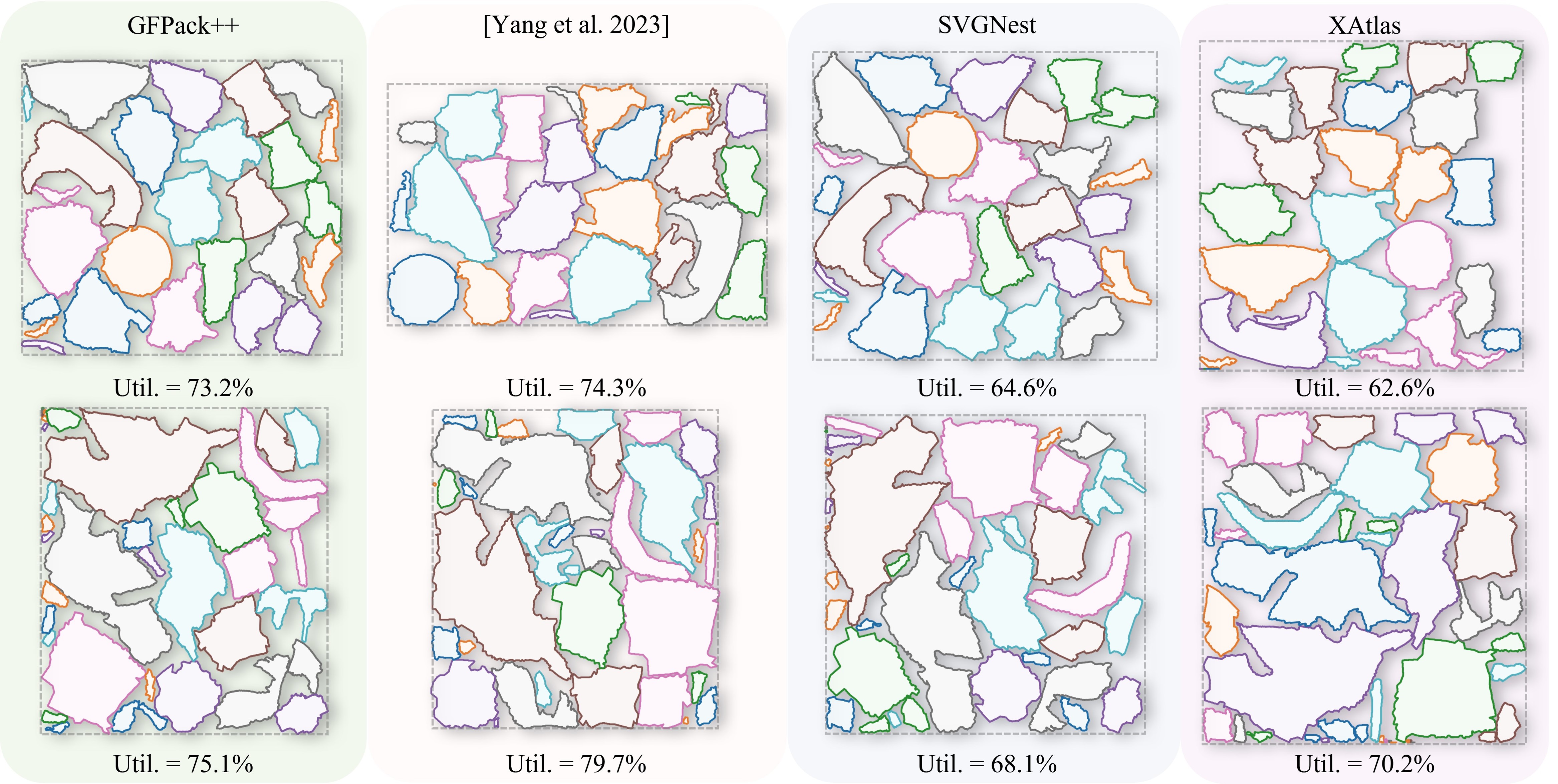}
\caption{Comparative results from {\gfpackplus}, \cite{Yang2023RLPack}, SVGnest, and XAtlas applied to Atlas datasets. 
}
\label{supp:fig:compare2learn2pack}
\end{figure*}

We visualized comparative results from {\gfpackplus}, SVGnest, and XAtlas applied to Garment, Dental, Puzzle, and Atlas datasets (Fig.~\ref{supp:fig:compare2XAtlas}).
We also visualized comparative results from {\gfpackplus}, \cite{Yang2023RLPack}, SVGnest, and XAtlas applied to Atlas datasets (Fig.~\ref{supp:fig:compare2learn2pack}). The results of \cite{Yang2023RLPack} are taken directly from the paper.

\end{document}